\documentclass{article}

 \usepackage[preprint]{neurips_2026}

\usepackage[utf8]{inputenc} 
\usepackage[T1]{fontenc}    
\usepackage{hyperref}       
\usepackage{url}            
\usepackage{booktabs}       
\usepackage{amsfonts}       
\usepackage{nicefrac}       
\usepackage{microtype}      
\usepackage{xcolor}         
\usepackage{amsmath}
\usepackage{graphicx}

\title{StruMPL: Multi-task Dense Regression under Disjoint Partial Supervision and MNAR Labels}

\author{%
  Reza M. Asiyabi\thanks{https://reza-asiyabi.github.io/} \\
  School of Geosciences, University of Edinburgh, UK \\
  National Centre for Earth Observation (NCEO), UK \\
  \texttt{reza.asiyabi@ed.ac.uk} \\
  \And
  Juan Alberto Molina-Valero \\
  Department of Spatial Sciences, Faculty of Environmental Sciences \\
  Czech University of Life Sciences Prague, Praha, Czech Republic \\
  \texttt{molina\_valero@fld.czu.cz} \\
  \And
  The SEOSAW Partnership \\
  School of Geosciences, University of Edinburgh, UK \\
  \And
  Steven Hancock \\
  School of Geosciences, University of Edinburgh, UK \\
  National Centre for Earth Observation (NCEO), UK \\
  \texttt{steven.hancock@ed.ac.uk} \\
  \And
  Casey M. Ryan \\
  School of Geosciences, University of Edinburgh, UK \\
  \texttt{casey.ryan@ed.ac.uk} \\
}

\begin{document}

\maketitle


\begin{abstract}
Estimating forest aboveground biomass (AGB) from Earth observation combines two structurally incompatible label sources: spaceborne lidar provides canopy structure at millions of locations but no biomass estimate, and ground-based plots provide biomass at thousands of biased locations but no metrics of structure. No single training sample carries labels for all target variables, plot labels are missing not at random (MNAR), and biomass is linked to the structural variables by known but biome-specific allometric laws. We formalise this as multi-task dense regression under heterogeneous disjoint partial supervision with MNAR labels and inter-task physical constraints, and propose StruMPL to address it jointly. A shared encoder feeds per-variable regression, imputation, and propensity heads for spatial MNAR correction, and a learnable physics module that evaluates the inter-task constraint on the model's own predictions at every pixel. The supervised loss uses an Augmented IPW (AIPW) pseudo-outcome with stop-gradients on the propensity and on the imputation baseline; we show analytically and empirically that both are necessary for joint optimisation to recover IPW-weighted stationary points while keeping the loss bounded. On two ecologically distinct biomes, StruMPL outperforms ablation variants and the closest published method on AGB RMSE and bias, with a stratified analysis showing AIPW reduces high-AGB bias by $\sim$54\%.
\end{abstract}

\section{Introduction}
\label{sec:intro}

Aboveground biomass (AGB) is a major carbon stock in forests~\citep{requena2019estimating} and a required input for national greenhouse gas inventories and carbon offset markets~\citep{ipcc2019refinement}. Measuring AGB directly is destructive and infeasible at scale, so field ecologists measure structural attributes (e.g. height, cover and tree diameter) and estimate AGB through regionally calibrated allometric equations~\citep{chave2014improved, feldpausch2012tree}. Earth observation (EO) makes it possible to extend this indirect measurement chain to entire continents: spaceborne lidar (NASA's GEDI~\citep{dubayah2020global}) provides canopy height and cover at millions of locations, and ground-based forest inventories provide AGB along with stem density and (sometimes) wood density~\citep{seosaw2021network, ifn4_spain_2024}.

A model that jointly predicts \emph{all} forest variables from the same input is desirable as: firstly, it would allow the structural and biomass predictions share a representation, and secondly it can produce maps that are bio-physically consistent with the allometric law that connects them. The two label sources, however, are structurally incompatible. GEDI footprints carry height and cover but no AGB; field plots carry AGB but no canopy structure at GEDI's scale. \emph{No single training sample has labels for all target variables}, and the source providing the most important target (AGB) accounts for less than 1\% of the training data. Worse, field plots are not a random sample of the forest population: surveyors avoid inaccessible terrain and degraded stands, biasing observed AGB toward intermediate values. Standard multi-task and semi-supervised regression methods do not address this combination of structural label-partitioning, severe imbalance, and missingness-not-at-random (MNAR).

We argue that large-scale forest attribute estimation is, at the level we attack here, a machine-learning problem with three coupled components. (i) The target variables must be predicted jointly, but their training labels come from disjoint sources covering disjoint subsets of the variable space, a condition stronger than standard partial-label learning. (ii) The plot labels are MNAR~\citep{rubin1976inference}: a model fitted to the observed distribution will systematically underestimate AGB in dense forests. (iii) The targets are linked by an allometric relationship $\text{AGB} \approx g(H, C, \text{SD}, \text{WD}; \boldsymbol{\phi})$ whose parameters $\boldsymbol{\phi}$ vary by biome and cannot be fitted directly from any sample, since no sample contains all variables together. These characteristics should be treated jointly: balancing supervision sources in batches without correcting for MNAR amplifies the bias; correcting for MNAR without exploiting cross-variable structure leaves the unlabelled majority of pixels without supervision.

We present \textbf{StruMPL} (Structured Multi-task Physics-constrained Learning), a single-objective framework that addresses the three problems jointly. A shared encoder produces features used by per-variable regression and imputation heads and by a propensity head that estimates per-pixel, per-variable observation probabilities. The regression objective is the Augmented Inverse Probability Weighting (AIPW) estimator~\citep{robins1994estimation, scharfstein1999adjusting}, with a stop-gradient design that lets the encoder be trained jointly with the propensity and imputation models without collapse. A differentiable physics module, with learnable allometric coefficients, evaluates the ecological law on the model's own predictions at every pixel, providing semi-supervised gradient at the >99\% of pixels that have no label. We evaluate the framework at two ecologically distinct sites: Mediterranean forests in Spain (SNFI~\citep{ifn4_spain_2024}), and dry tropical/savanna forests in Africa (SEOSAW~\citep{seosaw2021network}), demonstrating the framework's portability across biomes and across differing variable availability.

\paragraph{Contributions}
\begin{itemize}
    \item We formalize \emph{multi-task dense regression under heterogeneous disjoint partial supervision with MNAR labels and inter-task physical constraints}, a problem class appearing whenever structured domain knowledge links target variables observable only through separate, biased instruments. We test the formulation on AGB estimation across two biomes with different variable availability, using identical architecture and training recipe.
    \item We introduce a joint propensity model for spatial MNAR correction in dense regression and adopt the AIPW pseudo-outcome estimator as the supervised loss, with separate imputation heads as the outcome baseline via stop-gradient. We show analytically and empirically that two stop-gradients (on the propensity and imputation baseline) are individually necessary for stable joint training: the propensity detach prevents propensity collapse, while the imputation detach preserves the independence between outcome and imputation estimates that gives the AIPW pseudo-outcome its variance-reduction property over standard IPW.
    \item We design a differentiable, parametrised constraint module that evaluates a known inter-task biophysical relationship on the model's own predictions at every pixel, supplying a cross-source semi-supervised signal at locations with no label. Its parameters are learned jointly with the network, and its output is initialised to a domain-plausible range to avoid the gradient pathologies that otherwise prevent joint training. We represent it as a learnable allometric form for forest biomass.
    \item We validate the framework on two structurally distinct biomes (Mediterranean and dry tropical/savanna forests, with different target-variable availability) using identical architecture and training recipe. Stratified analysis shows the AIPW correction reduces bias specifically in the under-represented regime where MNAR effects concentrate; two distinct sites show that the unified mask design adapts cleanly when a target variable is missing.
\end{itemize}

\section{Related Work}
\label{sec:related}

\paragraph{Forest attribute estimation from EO.}
AGB and canopy variables have been estimated from satellite imagery using radar empirical models~\citep{ulaby1990michigan} and, more recently, deep networks~\citep{lang2023high, asiyabi2026process, weber2025unified}. Existing pipelines treat each variable as a separate target with independent training, or use disjoint per-source heads even when multiple variables are predicted~\citep{asiyabi2026process, guo2023combining}. But none enforce the ecological constraint that links the variables, and none address the MNAR structure of plot labels.

\paragraph{Multi-task learning under partial and heterogeneous supervision.}
Multi-task learning exploits inter-task statistical dependence to improve generalisation~\citep{caruana1997multitask}, with task selection and gradient interference being central concerns~\citep{standley2020tasks, yu2020gradient}. Partial-label MTL allows any subset of labels per sample~\citep{nishi2024joint}, and multi-source learning combines different supervision modalities~\citep{ouyang2014multi, hur2023genhpf, fifty2021efficiently}. Both approaches assume a common label inventory across sources or random missingness; neither addresses the case where source identity \emph{determines} which variables are labelled and the source distributions of the variables are linked by a known physical law.

\paragraph{MNAR and propensity-based correction.}
The MCAR/MAR/MNAR are introduced in~\citet{rubin1976inference} and subsequently developed by \citet{little2019statistical}. IPW~\citep{horvitz1952generalization, rosenbaum1983central} and AIPW~\citep{robins1994estimation, scharfstein1999adjusting} are standard tools for unbiased estimation under MAR; the latter has a doubly-robust property under classical conditions. These estimators have been used in causal inference~\citep{chernozhukov2018double}, recommendation~\citep{schnabel2016recommendations, saito2020unbiased}, almost always at the sample level on tabular covariates. We apply AIPW at \emph{pixel} resolution, with a per-pixel per-variable propensity estimated jointly with regression from a shared encoder, and characterise the gradient-separation needed for joint training.

\paragraph{Physics-informed and semi-supervised learning.}
Physics-Informed Neural Networks~\citep{raissi2019physics} encode physical laws as residual losses on a single target, typically with fixed parameters, and have been applied in fluid dynamics~\citep{cai2021physics}, climate~\citep{kashinath2021physics}, and ecology~\citep{viet2024adapting, miranda2024exploring}. Consistency regularisation~\citep{tarvainen2017mean, sohn2020fixmatch} provides another semi-supervised signal in dense prediction~\citep{ouali2020semi, ji2019semi}. Our physics module differs in two respects: its parameters are \emph{learned} jointly with the network, so the model is not committed to published exponents that may not transfer; and it links \emph{disjoint} label sources, providing a per-pixel pseudo-target derived from the model's own structural predictions rather than from a teacher network or a confident-prediction threshold.

\section{Problem Formulation}
\label{sec:problem}

\paragraph{Setup.}
Let $\mathbf{x} \in \mathbb{R}^{C_{in} \times H \times W}$ be a multi-channel EO patch, $\mathbf{y} \in \mathbb{R}^{K \times H \times W}$ a joint target tensor over $K$ forest attributes, and $R \in \{0,1\}^{K \times H \times W}$ a binary observation mask with $R_{k,i,j} = 1$ if variable $k$ is labelled at pixel $(i,j)$. The goal is a mapping $f_\theta: \mathbf{x} \mapsto \hat{\mathbf{y}}$ that minimises expected error over all variables and pixels in the target population.

\paragraph{Heterogeneous disjoint and sparse label structure.}
Training data is sparse (only a few labelled pixels in each patch) and comes from two sources. Spaceborne lidar (GEDI) provides $\mathcal{K}_G \subset \{1,\ldots,K\}$ at sparse $\sim$25\,m footprints, with $N_G$ samples; field plots provide $\mathcal{K}_P \subset \{1,\ldots,K\}$ at plot locations, with $N_P$ samples. By construction $\mathcal{K}_G \cap \mathcal{K}_P = \emptyset$: GEDI carries $\{H, C\}$ and plots carry $\{\text{SD}, \text{AGB}, \text{WD}\}$; the source determines which variables are labelled. The full training set is $\mathcal{D} = \mathcal{D}_G \cup \mathcal{D}_P$, with severe imbalance $N_G \gg N_P$ ($\sim$250:1 for Africa, $\sim$110:1 for Spain). This is stronger than standard partial-label learning, where the missingness is not deterministic by the source.

\paragraph{MNAR labels.}
Within $\mathcal{D}_P$, the labelled locations are not a random sample. Surveyors visit accessible forests at moderate elevation and avoid both inaccessible old-growth and degraded stands; the high-AGB tail is therefore underrepresented. Following~\citet{rubin1976inference}, labels are \emph{Missing Not At Random} (MNAR) when $P(R_{k,i,j} = 1 \mid \mathbf{x}, \mathbf{y})$ depends on $y_{k,i,j}$ even given $\mathbf{x}$. A model trained by the standard naive masked supervised loss optimises performance on the observed distribution rather than the population, and in our setting will systematically underestimate dense forests and overestimate degraded ones.

The standard correction is to model the propensity $\pi_{k,i,j} = P(R_{k,i,j}=1 \mid \mathbf{x})$ and re-weight by $1/\pi_{k,i,j}$~\citep{rosenbaum1983central}. This recovers an unbiased estimator under \emph{conditional ignorability} (i.e., that label availability depends on $\mathbf{x}$ but not on $y_{k,i,j}$ given $\mathbf{x}$). In our setting, however, ignorability does not hold exactly as the physical inaccessibility may correlate with biomass beyond what the EO data show at the patch scale. Our assumption is that the encoder's learned representation is a sufficiently rich covariate summary for ignorability to hold approximately. To reduce sensitivity to propensity misspecification we use the AIPW estimator~\citep{robins1994estimation}, which has a doubly-robust property under our working assumption of approximate ignorability: it is consistent if \emph{either} the propensity or the imputation model is correctly specified.

\paragraph{Physical constraint.}
Forest ecology gives an allometric relationship
\begin{equation}
    \text{AGB} \;\approx\; g\bigl(H,\, C,\, \text{SD},\, \text{WD};\, \boldsymbol{\phi}\bigr),
    \label{eq:allometry}
\end{equation}
with biome-specific parameters $\boldsymbol{\phi}$. Crucially, equation~\eqref{eq:allometry} \emph{cannot be fitted from the training data alone} as $\mathcal{K}_G \cap \mathcal{K}_P = \emptyset$ (i.e., $H, C$ are never co-observed with AGB, SD, WD on any sample). We therefore introduce $g$ as an external structural prior. Once $f_\theta$ produces $\hat{\mathbf{y}}$, the allometric residual $\hat{y}_\text{AGB} - g(\hat{y}_H, \hat{y}_C, \hat{y}_\text{SD}, \hat{y}_\text{WD};\boldsymbol{\phi})$ is defined at every pixel, labelled or not. This makes the physical constraint a natural cross-source semi-supervised signal at the >99\% of pixels with no label.

\paragraph{Multi-site variable availability.}
Survey protocols differ in which attributes they record: SEOSAW (Africa) reports $\mathcal{K}_P = \{\text{SD}, \text{WD}, \text{AGB}\}$ while SNFI (Spain) lacks wood density ($\mathcal{K}_P = \{\text{SD}, \text{AGB}\}$). Variable absence is encoded as an all-zero mask column, so the architecture is unchanged across sites; only the input set of the physics function $g$ changes. The sites also differ in test-set size: SNFI provides $\sim$6{,}000 plots, supporting stratified bias analysis; SEOSAW provides 162 standardised 1\,ha plots recommended by the data providers as the network's higher-quality subset (Appendix~\ref{app:Source-balanced sampling} and~\ref{app:portability}). Stratified analysis is therefore reported only on SNFI; aggregate metrics are reported on both.

\section{Method}
\label{sec:method}

\begin{figure}[t]
    \centering
    \includegraphics[width=1\textwidth]{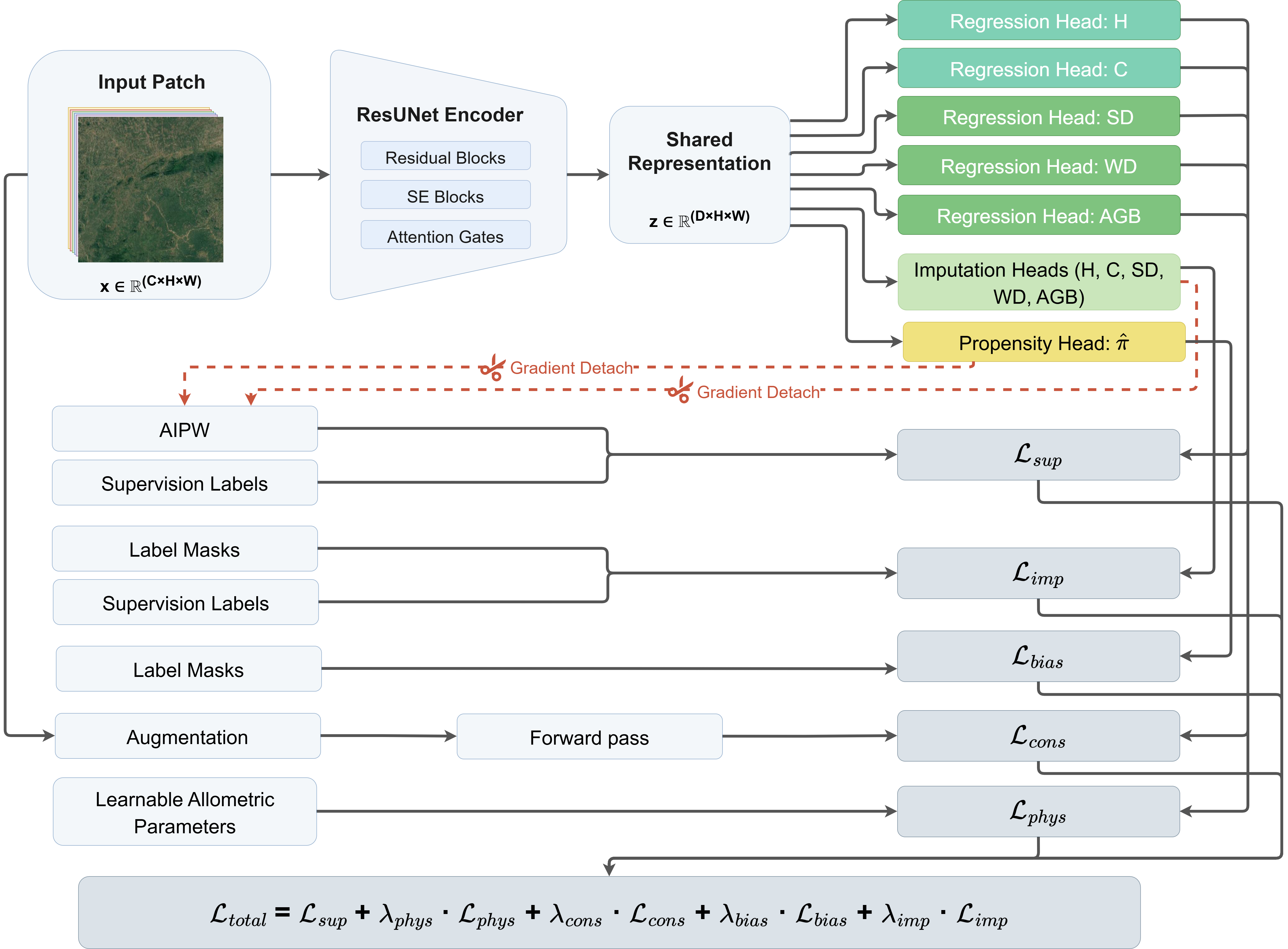}
    \caption{Overview of StruMPL. A shared encoder feeds regression, imputation and propensity heads. $\mathcal{L}_\text{sup}$ uses the AIPW pseudo-outcome with stop-gradients on $\hat{\boldsymbol{\pi}}$ and $\hat{m}$ (section~\ref{sec:lsup}); $\mathcal{L}_\text{imp}$ trains the imputation heads on the labelled pixels; $\mathcal{L}_\text{phys}$ enforces allometric consistency at every pixel via the learnable module $g_{\boldsymbol{\phi}}$; $\mathcal{L}_\text{cons}$ regularises under input augmentation; $\mathcal{L}_\text{bias}$ trains $\hat{\boldsymbol{\pi}}$ against the mask.}
    \label{fig:architecture}
\end{figure}

StruMPL addresses the three problems above within a single forward pass and a unified objective (Figure~\ref{fig:architecture}). A shared encoder $\text{Enc}_\theta : \mathbf{x} \mapsto \mathbf{z} \in \mathbb{R}^{D \times H \times W}$ produces features used by $K$ independent regression heads (each producing $\hat{y}_k \in \mathbb{R}^{H \times W}$ in z-score space), $K$ independent imputation heads (each producing $\hat{m}_k \in \mathbb{R}^{H \times W}$ in z-score space) and a propensity head $\hat{\boldsymbol{\pi}} = \sigma(\text{BiasHead}(\mathbf{z})) \in (0,1)^{K \times H \times W}$. We use a ResUNet with channel and spatial attention as the encoder; full architectural details are in Appendix~\ref{app:arch}.

The total objective is
\begin{equation}
    \mathcal{L}_\text{total} \;=\; \mathcal{L}_\text{sup} \;+\; \lambda_\text{phys}\, \mathcal{L}_\text{phys}
    \;+\; \lambda_\text{cons}\, \mathcal{L}_\text{cons} \;+\; \lambda_\text{bias}\, \mathcal{L}_\text{bias} \;+\; \lambda_\text{imp}\, \mathcal{L}_\text{imp},
    \label{eq:total_loss}
\end{equation}
with each term targeting one of the failure modes identified in section~\ref{sec:problem}. We describe the two terms central to the contribution ($\mathcal{L}_\text{sup}$ and $\mathcal{L}_\text{phys}$) below; $\mathcal{L}_\text{cons}$, $\mathcal{L}_\text{bias}$ and $\mathcal{L}_\text{imp}$ are standard and we summarise them in section~\ref{sec:other_losses} (full forms in Appendix~\ref{app:losses}).

\subsection{Supervised loss with AIPW MNAR correction}
\label{sec:lsup}
The naive masked MSE optimises performance on the observed distribution. We replace it with the AIPW pseudo-outcome estimator. Let $\mu_{k,i,j} = \text{sg}(\hat{m}_{k,i,j})$ denote the imputation estimate with gradients stopped, and $\hat{\pi}_{k,i,j}$ the propensity (also gradients stopped), clamped to $[\pi_\text{min}, 1]$ with $\pi_\text{min}=0.1$ to bound the effective AIPW weight. We define the pseudo-outcome
\begin{equation}
    \tilde{y}_{k,i,j} \;=\; \mu_{k,i,j} \;+\; \frac{R_{k,i,j}}{\hat{\pi}_{k,i,j}} \bigl( y_{k,i,j} - \mu_{k,i,j} \bigr).
    \label{eq:dr_pseudooutcome}
\end{equation}
The supervised loss is a pixel-wise MSE between the regression output and the pseudo-outcome:
\begin{equation}
    \mathcal{L}_\text{sup} \;=\; \frac{1}{K \cdot H W}
    \sum_{k,i,j} \bigl( \hat{y}_{k,i,j} - \tilde{y}_{k,i,j} \bigr)^2.
    \label{eq:sup_loss}
\end{equation}

\paragraph{Stop-gradients are not optional.}
Two stop-gradients in Eq.~\eqref{eq:dr_pseudooutcome} keep joint training stable. Without detaching $\hat{\pi}$, the encoder could minimise $\mathcal{L}_\text{sup}$ by inflating $\hat{\pi} \to 1$, collapsing the IPW weight and erasing the MNAR correction. Without detaching $\mu$, $\mathcal{L}_\text{sup}$ would backpropagate into the imputation heads and drive $\hat{m} \to \hat{y}$, collapsing $\mathcal{L}_\text{sup}$ to a $1/\hat{\pi}^2$-weighted MSE on labelled pixels (worse than standard IPW). The propensity and imputation heads are therefore trained only through $\mathcal{L}_\text{bias}$ and $\mathcal{L}_\text{imp}$ (section~\ref{sec:other_losses}), and enter $\mathcal{L}_\text{sup}$ as fixed coefficients. With both stop-gradients applied, the gradient w.r.t.\ $\hat{y}$ is $2(\hat{y} - \tilde{y})$, with no gradient flowing into $\hat{m}$ or $\hat{\pi}$. At unlabelled pixels ($R=0$), this pulls $\hat{y}$ toward $\mu$; at labelled pixels ($R=1$), the pseudo-outcome combines $\mu$ with an IPW-corrected residual, so the model is pushed toward the AIPW-weighted stationary points. When $\mu$ is reasonably well-fit, the residual $(y - \mu)$ is small, reducing the $O(1/\hat{\pi})$ loss spikes characteristic of standard IPW; this is the variance-reduction property of AIPW over IPW (full derivation in Appendix~\ref{app:gradient}).

We note one caveat: classical double robustness requires cross-fitting for asymptotic independence of the estimators \citep{chernozhukov2018double}. Due to our shared-encoder architecture, we use AIPW for empirical robustness to either-side misspecification rather than as a formal consistency guarantee.

\subsection{Physics consistency loss}
\label{sec:lphys}

The physics module $g_{\boldsymbol{\phi}}$ is a learnable allometric equation evaluated on the model's own structural predictions. We use the ecology-motivated form
\begin{equation}
    \widehat{\text{AGB}}_\text{phys} \;=\; \alpha \;+\; \text{scale} \cdot
    \bigl[\,\text{sp}(\hat{y}_H)^b \cdot \text{sp}(\hat{y}_C)^c \bigr]^{\,\text{sp}(\hat{y}_\text{SD}) \cdot d}
    \cdot \text{sp}(\hat{y}_\text{WD})^{e},
    \label{eq:allometric_phys}
\end{equation}
where $\boldsymbol{\phi} = \{\alpha, \text{scale}, b, c, d, e\}$ are constrained positive via softplus ($\text{sp}(\cdot) = \log(1 + \exp(\cdot))$) or exp. The interaction term $[\text{sp}(\hat{y}_H)^b \cdot \text{sp}(\hat{y}_C)^c]^{\text{sp}(\hat{y}_\text{SD}) \cdot d}$ allows stem density to modulate the structural-to-biomass scaling, motivated by competition effects in dense stands. When \text{WD} is unavailable (Spain), the $\text{sp}(\hat{y}_\text{WD})^e$ factor is dropped and $|\boldsymbol{\phi}|$ decreases by one. The form is an ecologically-motivated parametrisation rather than a derivation; we compare it against a plain multiplicative power law and a learned MLP in section~\ref{sec:results}.

The regression heads operate in z-score space; the physics module operates in physical units. We therefore denormalise the structural predictions before equation~\eqref{eq:allometric_phys}, clamp the output to the ecologically plausible range $[0, 2000]$\,Mg/ha, and renormalise to z-score space using the training-set statistics. The physics consistency loss is then a pixelwise MSE applied to all pixels, labelled or not:
\begin{equation}
    \mathcal{L}_\text{phys} \;=\; \frac{1}{HW} \sum_{i,j}
    \bigl( \hat{y}_{\text{AGB},i,j} - \widehat{\text{AGB}}^{\text{norm}}_{\text{phys},i,j} \bigr)^2.
    \label{eq:phys_loss}
\end{equation}
Gradients flow from $\mathcal{L}_\text{phys}$ back into all structural heads, into the AGB head, and into $\boldsymbol{\phi}$. This is the mechanism by which the two label sources exchange information: GEDI-supervised $\hat{y}_H, \hat{y}_C$ are constrained, under the learned allometry, to be consistent with plot-supervised $\hat{y}_\text{AGB}$.

\subsection{Other components}
\label{sec:other_losses}

\paragraph{Propensity training.}
The BiasHead is trained via binary cross-entropy between $\hat{\boldsymbol{\pi}}$ and the observation mask $R$, applied at every pixel. Gradients flow back through the shared encoder, so the encoder is trained to represent features predictive of survey accessibility, i.e., exactly the features the propensity model needs. The full form is in Appendix~\ref{app:losses}.

\paragraph{Augmentation consistency.}
We add a standard FixMatch-style consistency term~\citep{tarvainen2017mean, sohn2020fixmatch}: different predictions on $\mathbf{x}$ and on a perturbed view $\mathbf{x}'$ (additive Gaussian noise and per-element channel dropout) are penalised. We do not use EMA teacher and confidence threshold, as the small patch size and the stronger physics signal make these unnecessary.

\paragraph{Imputation training.}
The imputation heads $\hat{m}_k$ are trained via masked MSE on labelled pixels only, providing the outcome baseline used in the AIPW pseudo-outcome (section~\ref{sec:lsup}). They are deliberately trained on the (biased) observed distribution: this is what AIPW expects of the outcome model, and the propensity correction in $\mathcal{L}_\text{sup}$ debiases the final regression heads.

\paragraph{Implementation details.}
Complete implementation details including architecture components, hyperparameters, source-balanced batching, parameters initialisation, and parameter count are given in Appendix~\ref{app:arch},~\ref{app:Source-balanced sampling}, and~\ref{app:impl}.

\section{Results and Discussion}
\label{sec:results}

We evaluate StruMPL on two datasets covering ecologically distinct biomes: \textbf{Spain} (SNFI, $\sim$6{,}000 test plots) and \textbf{Africa} (SEOSAW, 162 test plots in the standardised 1\,ha format). For headline numbers we report mean and standard deviation over 5 random seeds (42, 123, 456, 789, 1011); ablation rows use a single seed (42) to keep compute tractable.

\subsection{Headline performance and the cost of each component}
\label{sec:headline}

\begin{figure}[t]
    \centering
    \includegraphics[width=1\textwidth]{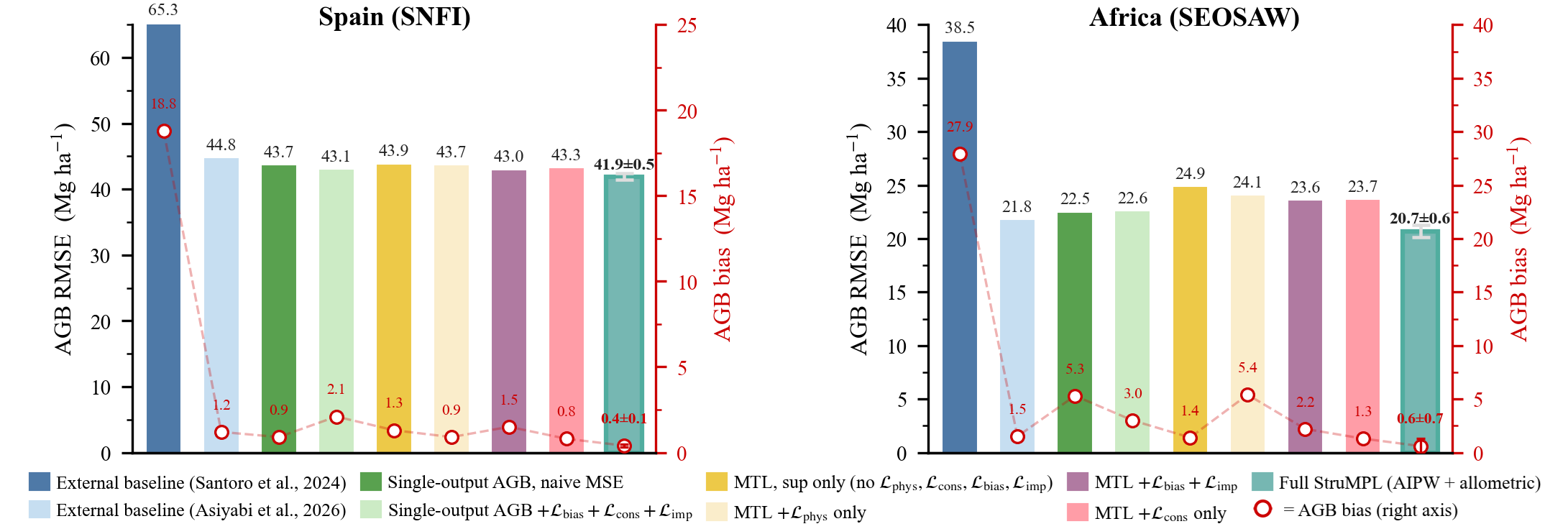}
    \caption{AGB RMSE (bars, left axis) and AGB bias (red markers, right axis) for all model configurations from Table~\ref{tab:main} on the Spain (left) and Africa (right) test sets. Lower is better for both metrics. Error bars on Full StruMPL denote standard deviation across 5 random seeds.}
    \label{fig:RMSE_Bias}
\end{figure}

Table~\ref{tab:main} and Figure~\ref{fig:RMSE_Bias} report AGB RMSE and bias for full StruMPL, MTL variants with loss-component ablations, two single-output baselines, and two published external baselines (\citep{asiyabi2026process} closest to StruMPL, and \citep{santoro2024biomasscci} standard global AGB map from ESA CCI; see Appendix~\ref{app:external}). The complete ablation grid and per-variable RMSE for the structural variables are in Appendix~\ref{app:full_ablation_table}; here we focus on AGB as the main target variable.

\begin{table}[t]
\centering
\footnotesize
\caption{Main results on the primary AGB target (Mg/ha). Lower is better for both RMSE and $|\text{bias}|$. Full StruMPL reports mean $\pm$ std over 5 seeds; ablations are single-seed.}

\label{tab:main}
\begin{tabular}{lccccc}
\toprule
& \multicolumn{2}{c}{Spain (SNFI)} & \multicolumn{2}{c}{Africa (SEOSAW)} \\
\cmidrule(lr){2-3} \cmidrule(lr){4-5}
Configuration & AGB RMSE & AGB bias & AGB RMSE & AGB bias \\
\midrule
External baseline (\citep{santoro2024biomasscci}) & \texttt{65.3} & \texttt{18.8} & \texttt{38.5} & \texttt{27.9} \\
External baseline (\citep{asiyabi2026process}) & \texttt{44.8} & \texttt{1.2} & \texttt{21.8} & \texttt{1.5} \\
Single-output AGB, naive MSE & \texttt{43.7} & \texttt{0.9} & \texttt{22.5} & \texttt{5.3} \\
Single-output AGB + $\mathcal{L}_\text{bias}$ + $\mathcal{L}_\text{cons}$ + $\mathcal{L}_\text{imp}$ & \texttt{43.1} & \texttt{2.1} & \texttt{22.6} & \texttt{3.0} \\
\midrule
\multicolumn{5}{l}{\emph{Joint multi-task variants}} \\
MTL, sup only (no $\mathcal{L}_\text{phys}, \mathcal{L}_\text{cons}, \mathcal{L}_\text{bias}, \mathcal{L}_\text{imp}$) & \texttt{43.9} & \texttt{1.3} & \texttt{24.9} & \texttt{1.4} \\
MTL + $\mathcal{L}_\text{phys}$ only & \texttt{43.7} & \texttt{0.9} & \texttt{24.1} & \texttt{5.4} \\
MTL + $\mathcal{L}_\text{bias}$ + $\mathcal{L}_\text{imp}$ & \texttt{43.0} & \texttt{1.5} & \texttt{23.6} & \texttt{2.2} \\
MTL + $\mathcal{L}_\text{cons}$ only & \texttt{43.3} & \texttt{0.8} & \texttt{23.7} & \texttt{1.3} \\
\textbf{Full StruMPL (AIPW + allometric)} & \textbf{\texttt{41.9 $\pm$ 0.5}} & \textbf{\texttt{0.4 $\pm$ 0.1}} & \textbf{\texttt{20.7 $\pm$ 0.6}} & \textbf{\texttt{0.6 $\pm$ 0.7}} \\
\bottomrule
\end{tabular}
\end{table}

\paragraph{Discussion.}
Full StruMPL achieves the best AGB RMSE and bias on both sites, but the margins tell different stories. On Africa, where the source imbalance is larger, StruMPL shows a clear gain over every ablation: 20.7 vs.\ 22.5 Mg/ha against the strongest single-component variant, and 24.9 Mg/ha against the MTL baseline with no propensity correction or physics consistency. The MTL-only configuration is in fact the worst row on Africa, showing that MTL without the StruMPL components is actively harmful, and that the observed gains are not a generic side-effect of multi-tasking but specifically attributable to $\mathcal{L}_\text{phys}$ and $\mathcal{L}_\text{bias}$. The Africa results also support the portability claim: the unified mask design adapts cleanly to wood density availability, the \textit{BalancedBatchSampler} handles the larger source imbalance ($N_G/N_P \approx 250$ vs.\ Spain's $\approx 110$), and the curriculum warmup is effective without recipe changes (further discussed in Appendix~\ref{app:portability}).

On Spain, the gain over the strongest ablation is smaller but statistically robust: full StruMPL reaches 41.9 $\pm$ 0.5 Mg/ha against 43.0--43.9 for the MTL variants and 43.1 for the strongest single-output baseline. The improvement over the strongest ablation (MTL + $\mathcal{L}_\text{bias} + \mathcal{L}_\text{imp}$) is significant under a paired bootstrap ($p < 0.05$, $n = 10{,}000$; Appendix~\ref{app:bootstrap}). The clearer StruMPL signal on Spain is in bias: full StruMPL achieves $|\text{bias}| = 0.4 \pm 0.1$\,Mg/ha against 0.8--2.1 for the ablations, which is the impact of the AIPW correction (section~\ref{sec:stratified}). StruMPL also outperforms the closest published method~\citep{asiyabi2026process} on both sites in both metrics, and is substantially more accurate than the off-the-shelf ESA CCI global biomass map~\citep{santoro2024biomasscci}, which we include as a reference point.

\subsection{Where AIPW pays off: stratified bias on Spain}
\label{sec:stratified}

\begin{figure}[h]
    \centering
    \includegraphics[width=0.8\textwidth]{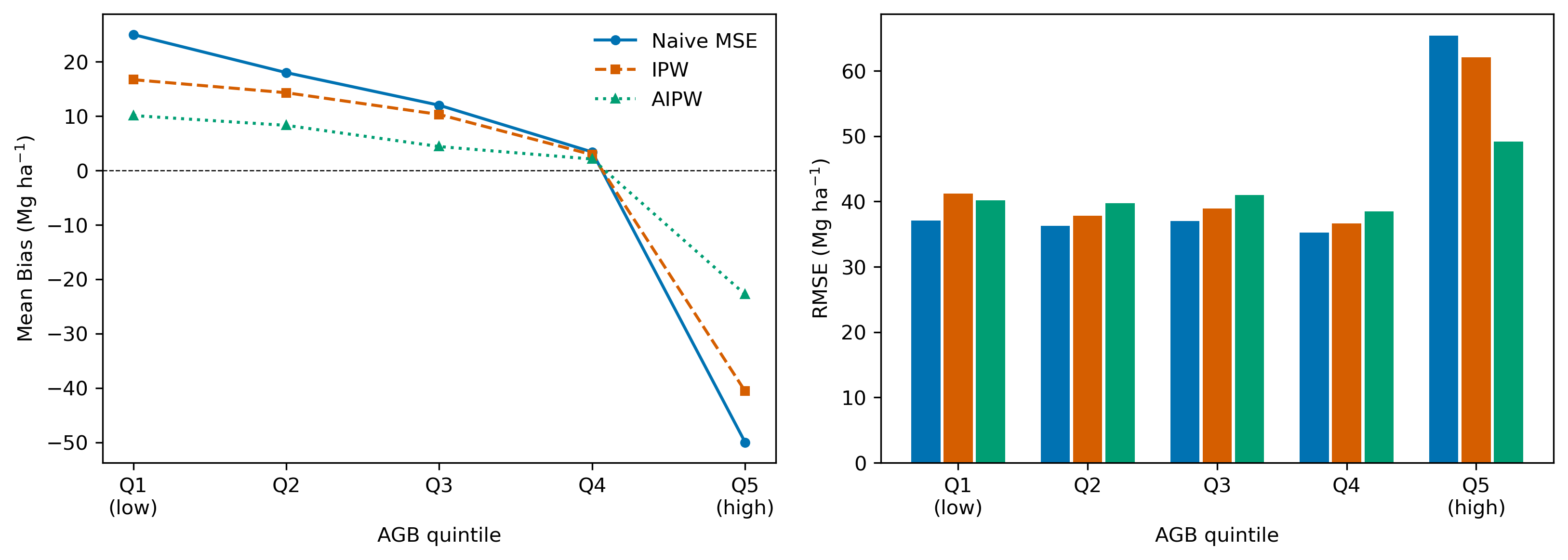}
    \caption{Bias and RMSE on AGB across five quantiles for naive masked MSE, IPW, and AIPW supervision (all other components fixed). AIPW reduces the systematic high-AGB underestimation that naive MSE exhibits; per-quantile RMSE redistributed toward the under-represented high tail.}
    \label{fig:stratified}
\end{figure}

The AIPW correction is designed to reduce systematic high-AGB underestimation where MNAR effects concentrate. Aggregate metrics hide this because the high-AGB tail is a small fraction of test samples. We bin Spain test pixels (due to the higher number of the test plots) into 5 AGB quantiles by plot label and report RMSE and bias per quantile (Figure~\ref{fig:stratified}).

\paragraph{Discussion.}
The naive masked MSE shows the predicted MNAR pattern: overestimation in low AGB regions ($+25$\,Mg/ha in Q1) and large underestimation in the high-AGB tail ($-50$\,Mg/ha in Q5), crossing zero around Q4; whole-dataset bias ($+1.7$\,Mg/ha) hides this entirely as the two ends cancel. IPW reduces Q5 underestimation modestly ($-40.6$\,Mg/ha) while AIPW reduces it substantially ($-22.8$\,Mg/ha, a 54\% reduction); lower quantiles improve proportionally (Q1: $+25 \to +10.1$). The RMSE pattern reflects the same redistribution: AIPW reduces Q5 RMSE from 65.4 to 49.2\,Mg/ha while marginally increasing RMSE in Q1--Q4 (by 3--4\,Mg/ha), as expected from reweighting toward under-represented samples. The net effect on whole-dataset RMSE is positive (43.2 to 41.9\,Mg/ha), and the framework prioritises the regime where MNAR error is concentrated.

\subsection{Are the stop-gradients necessary?}
\label{sec:detach_ablation}

In section~\ref{sec:lsup} and Appendix~\ref{app:gradient}, we mentioned that two stop-gradients in the AIPW pseudo-outcome are individually necessary for stable joint training. We test this empirically (Table~\ref{tab:detach}).

\begin{table}[h]
\centering
\small
\caption{Stop-gradient ablation on the Spain test set, full StruMPL configuration.}
\label{tab:detach}
\begin{tabular}{lccc}
\toprule
Configuration & AGB RMSE & Mean $\hat\pi$ at convergence & Outcome \\
\midrule
Both stop-gradients (default) & \texttt{41.9}& \texttt{0.62}& stable \\
$\hat\pi$ \emph{not} detached in $\mathcal{L}_\text{sup}$ & \texttt{44.2}& \texttt{0.99}& propensity collapse \\
$\mu = \hat y$ \emph{not} detached & \texttt{44.6}& \texttt{0.60}& imputation heads collapse \\
\bottomrule
\end{tabular}
\end{table}

\paragraph{Discussion.}
Without detaching $\hat\pi$, the encoder reduces $\mathcal{L}_\text{sup}$ by artificially inflating predicted propensities; we observe mean $\hat\pi$ drift to $\approx 1.0$, which collapses the AIPW correction into a naive, biased MSE. Without detaching $\mu$, the imputation heads lose their functional independence and begin to shadow the main regression heads. This results in the $1/\hat{\pi}^2$ weighting failure described in Appendix~\ref{app:gradient}, which amplifies noise and degrades performance. These empirical failures confirm that the stop-gradients are structural requirements of the AIPW framework.

\subsection{Ecologically motivated physics module}
\label{sec:phys_ablation}

In Table~\ref{tab:phys_form}, we compare the chosen \texttt{allometric} parametrisation against \texttt{power\_law} (a plain multiplicative power law, the conservative baseline) and \texttt{mlp} (a 3-layer MLP, an unconstrained baseline).

\begin{table}[h]
\centering
\small
\caption{Physics module form; full StruMPL otherwise.}
\label{tab:phys_form}
\begin{tabular}{lccc}
\toprule
Physics module & Spain AGB RMSE  & Africa AGB RMSE & Notes \\
\midrule
\texttt{power\_law} & \texttt{44.3} & \texttt{25.5} & baseline \\
\texttt{allometric} (default) & \texttt{41.9}&\texttt{20.7}& ecology motivated \\
\texttt{mlp} & \texttt{44.4}&\texttt{24.2}& not interpretable \\
\bottomrule
\end{tabular}
\end{table}

\paragraph{Discussion.}
The \texttt{allometric} form outperforms both alternatives on both sites, reducing AGB RMSE by 2.4\,Mg/ha (5.4\%) over \texttt{power\_law} on Spain and 4.8\,Mg/ha (18.8\%) on Africa. A 3-layer MLP has more parameters and more flexibility than either parametrised form, yet performs comparably to the plain \texttt{power\_law} on Spain (44.4 vs.\ 44.3) and remains worse than \texttt{allometric} on Africa. The improvement therefore comes from inductive bias rather than expressiveness. The relative gain is larger on Africa than on Spain. This observation is consistent with the ecological intuition that competition-driven scaling effects matter more in the heterogeneous-density savanna stands of the SEOSAW domain than in the more homogeneous Mediterranean stands of SNFI; we report this as a qualitative observation rather than a quantitative claim, given the smaller Africa test set. We use \texttt{allometric} as the default and use its learned coefficients in the cross-site comparison of section~\ref{sec:exponents}.

\subsection{Learned allometric coefficients}
\label{sec:exponents}

The allometric parameters $\boldsymbol{\phi} = \{\alpha, \text{scale}, b, c, d, e\}$ are jointly trained via $\mathcal{L}_\text{phys}$. Their fitted values are not maximum-likelihood ecological estimates as they are optimised under the full StruMPL objective (including AIPW reweighting and consistency regularisation), but they characterise the effective satellite-scale scaling that the model converges to. Table~\ref{tab:exponents} reports them across sites.

\begin{table}[h]
\centering
\small
\caption{Learned allometric exponents. Spain has no WD term ($e$ absent). Cross-site comparison is qualitative; see text.}
\label{tab:exponents}
\begin{tabular}{lcccccc}
\toprule
Site & $\alpha$ & scale & $b$ (H) & $c$ (C) & $d$ (SD) & $e$ (WD) \\
\midrule
Spain  & \texttt{0.82}& \texttt{31.3}& \texttt{0.04}& \texttt{0.21}& \texttt{0.02}& --- \\
Africa & \texttt{0.54}& \texttt{21.0}& \texttt{0.05}& \texttt{0.03}& \texttt{0.03}& \texttt{0.02}\\
\bottomrule
\end{tabular}
\end{table}

\paragraph{Discussion.}
Several points stand out in Table~\ref{tab:exponents}. First, the structural exponents ($b, c, d, e$) are uniformly small, far from the near-unit values of textbook allometric equations~\citep{chave2014improved}. We do not interpret this as a recovery of the textbook allometry: under the StruMPL objective, the optimiser finds a regime in which AGB is largely explained by the additive baseline $\alpha$ and the multiplicative $\text{scale}$, with the structural exponents acting as small corrections. This matches the framing of section~\ref{sec:lphys}: $\boldsymbol{\phi}$ are effective satellite-scale scaling coefficients, not estimates of the underlying ecological law. Second, the cross-site differences are concentrated in two parameters. The Spain $\text{scale}$ ($31.3$) exceeds Africa's ($21.0$), consistent with the higher mean AGB density of Mediterranean stands; and the canopy-cover exponent $c$ is markedly higher on Spain ($0.21$ vs.\ $0.03$), aligning with the intuition that denser-canopy Mediterranean forests show a stronger biomass-cover relationship than savanna stands, where local cover can be high (an isolated tree) without corresponding landscape biomass. The remaining exponents are roughly comparable across sites. We treat these readings as qualitative ecological observations rather than quantitative estimates.

\subsection{Limitations}
\label{sec:limitations}
StruMPL relies on two assumptions. First, conditional ignorability (i.e., the labels are MAR given the encoder's learned representation, i.e., $P(R=1 \mid z, y) = P(R=1 \mid z)$), however, in our setting the labels are MNAR with respect to the raw inputs $\mathbf{x}$ (plot accessibility correlates with biomass), and we rely on encoder to absorb enough of this dependence into $z$ that conditional independence holds approximately. AIPW does not protect against violations of this assumption itself; it protects only against misspecification of the propensity \emph{or} the imputation model when ignorability holds. Second, the formal double-robustness guarantee for ML estimators requires cross-fitting to control overfitting bias~\citep{chernozhukov2018double}. We share the encoder between the heads and do not cross-fit, so we use AIPW as an empirical bias-reduction tool rather than a formal consistency guarantee. Both points constrain the theoretical claims we can make but do not invalidate the empirical bias reduction reported in section~\ref{sec:stratified}. A separate caveat is the choice of allometric form: ours is one of several plausible parametrisations, and the learned exponents should be read as effective scaling coefficients under the StruMPL objective, not ecological estimates. We compare against alternative forms in section~\ref{sec:phys_ablation}.

\section{Conclusion}

We presented StruMPL, a framework for multi-task dense regression under disjoint heterogeneous partial supervision with MNAR labels and inter-task physical constraints. The framework combines a joint propensity model for spatial MNAR correction, an AIPW pseudo-outcome loss with two stop-gradients whose necessity we establish both analytically and empirically, and a learnable physics module that provides cross-source semi-supervised gradient at every pixel. Applied to forest attribute estimation on two ecologically distinct biomes, StruMPL outperforms single-output baselines, ablation variants, and the closest published method, with the AIPW correction specifically reducing bias in the high-AGB regime where MNAR effects are concentrated. The four-condition characterisation of when the formulation applies (Appendix~\ref{app:generality}) suggests that domains with structurally similar properties (hydrology, battery state estimation, multi-modal medical imaging) are candidate applications. We leave cross-domain test and addressing the limitations (section~\ref{sec:limitations}) to future work.

\newpage
\bibliographystyle{plainnat}
\bibliography{refs}

\newpage
\appendix
\section*{Appendix}
\addcontentsline{toc}{section}{Appendix}

\section{Architecture}
\label{app:arch}

\paragraph{Shared encoder.}
The encoder $\text{Enc}_\theta$ is a ResUNet~\citep{he2016deep} with residual blocks, Squeeze-and-Excitation channel attention~\citep{hu2018squeeze}, and spatial attention gates on the decoder skip connections~\citep{oktay2018attention}. Given $\mathbf{x} \in \mathbb{R}^{C_{in} \times H \times W}$ with $H = W = 16$, it produces $\mathbf{z} = \text{Enc}_\theta(\mathbf{x}) \in \mathbb{R}^{D \times H \times W}$ with $D = 128$. The encoder follows the standard U-Net contractive--expansive structure with 3 encoder stages, channel widths doubling at each downsampling level, and matching upsampling in the decoder. The shared encoder is motivated by two observations specific to this problem. First, $N_G \gg N_P$: borrowing statistical strength from the larger GEDI corpus is only possible under a shared representation. Second, the physics consistency loss links variables supervised by different sources, and the residual is meaningful only if the structural and biomass predictions are computed from a compatible feature space.

The input contains $C_{in} = 15$ channels drawn from a multi-sensor EO time series: ALOS PALSAR-2 L-band SAR backscatter~\citep{JAXA2021PALSAR2}, Harmonized Landsat Sentinel-2 (HLS) optical reflectance bands~\citep{masek2021hls}, and geographic coordinates. Channel attention lets the encoder learn input-channel importance per task; spatial attention on skip connections suppresses uninformative spatial regions, useful given that labels cover $<$1\% of pixels per patch.

\paragraph{Regression heads.}
$K$ independent two-layer convolutional heads map $\mathbf{z}$ to per-pixel predictions,
\begin{equation}
    \hat{y}_k = \text{Head}_k(\mathbf{z}), \qquad k \in \{1, \ldots, K\}.
\end{equation}
Each head consists of a $3 \times 3$ convolution ($D \to D/2$ channels) with RELU activation, followed by a $1 \times 1$ convolution ($D/2 \to 1$). We keep heads independent because each variable receives gradient from a different mask and a different source; sharing parameters would induce gradient interference between tasks with incompatible observability. Inter-variable correlation is enforced through $\mathcal{L}_\text{phys}$, which provides cross-variable signal without requiring shared head parameters.

All predictions are produced in z-score normalised space (zero mean, unit variance per variable, computed on the training set). This makes $\mathcal{L}_\text{sup}$ scale-invariant across variables with different physical ranges, makes $\mathcal{L}_\text{phys}$ numerically comparable to $\mathcal{L}_\text{sup}$, and keeps gradient magnitudes in a predictable range.

\paragraph{Imputation heads.}
Similar to the regression heads, $K$ independent two-layer convolutional heads with the same architecture ($3 \times 3$ conv $D \to D/2$ with RELU, followed by $1 \times 1$ conv $D/2 \to 1$) produce the per-pixel imputation outputs,
\begin{equation}
    \hat{m}_k = \text{ImputationHead}_k(\mathbf{z}), \qquad k \in \{1, \ldots, K\}.
\end{equation}
The imputation heads share the encoder with the regression and bias heads, but are kept architecturally separate from the regression heads, with no parameter sharing between $\hat{y}_k$ and $\hat{m}_k$. This separation preserves the functional independence between the outcome and imputation estimators that gives AIPW its variance-reduction property over standard IPW (section~\ref{sec:lsup}). Like the regression heads, imputation outputs are produced in z-score normalised space using the same training-set statistics, so $\mu_k = \text{sg}(\hat{m}_k)$ enters $\mathcal{L}_\text{sup}$ on the same scale as $\hat{y}_k$ and $y_k$. The imputation heads are trained exclusively through $\mathcal{L}_\text{imp}$ (Appendix~\ref{app:losses}); gradients from $\mathcal{L}_\text{sup}$ are stopped at $\hat{m}_k$, preventing the imputation model from collapsing onto the regression head (Appendix~\ref{app:gradient}).

\paragraph{Bias head.}
The BiasHead is a small convolutional head with sigmoid output,
\begin{equation}
    \hat{\boldsymbol{\pi}} = \sigma\!\left(\text{BiasHead}(\mathbf{z})\right) \in (0,1)^{K \times H \times W}.
\end{equation}
Its architecture mirrors the regression and imputation heads ($3 \times 3$ conv $D \to D/2$ with RELU, $1 \times 1$ conv $D/2 \to K$) so that the propensity estimation capacity matches the regression capacity. It shares the encoder with the regression heads, so the features that drive prediction also drive propensity estimation.

\paragraph{Parameter count.}
The full StruMPL model has $\sim$16.9\,M parameters, of which the encoder accounts for 16{,}204{,}198, the regression heads for 369{,}605 (73{,}921 each head), the imputation heads also for 369{,}605 (73{,}921 each head), the BiasHead for 74{,}181, and the physics module for 6 (a constant; the physics parameters are scalar). Removing the imputation heads, BiasHead and physics module gives a standard multi-task ResUNet.

\section{Full forms of $\mathcal{L}_\text{bias}$, $\mathcal{L}_\text{cons}$ and $\mathcal{L}_\text{imp}$}
\label{app:losses}

\paragraph{Propensity training loss.}
\begin{equation}
    \mathcal{L}_\text{bias} = -\frac{1}{K \cdot H W}
    \sum_{k,i,j} \bigl[\, R_{k,i,j} \log \hat{\pi}_{k,i,j} + (1 - R_{k,i,j}) \log (1 - \hat{\pi}_{k,i,j}) \bigr].
    \label{eq:bias_loss}
\end{equation}
This loss is evaluated at every pixel: labelled pixels are positive examples, unlabelled pixels are negative. Gradients flow back through the BiasHead into the encoder, training it to represent features correlated with label availability, exactly the features that the propensity model needs for downstream MNAR correction.

\paragraph{Augmentation consistency loss.}
\begin{equation}
    \mathcal{L}_\text{cons} = \frac{1}{H W} \sum_{i,j}
    \bigl\| \hat{\mathbf{y}}_{i,j}(\mathbf{x}) - \hat{\mathbf{y}}_{i,j}(\mathbf{x}') \bigr\|^2,
    \qquad
    \mathbf{x}' = \mathbf{x} \odot \mathbf{B} + \boldsymbol{\varepsilon},
\end{equation}
with $\boldsymbol{\varepsilon} \sim \mathcal{N}(\mathbf{0}, \sigma^2 \mathbf{I})$ ($\sigma = 0.05$) and $\mathbf{B} \sim \text{Bernoulli}(1 - p_\text{drop})^{C_{in} \times H \times W}$ ($p_\text{drop} = 0.05$). The perturbations are semantically neutral for EO patches at $16 \times 16$ resolution. We do not use Exponential Moving Average (EMA) teacher and confidence thresholding (cf.\ FixMatch~\citep{sohn2020fixmatch}, Mean Teacher~\citep{tarvainen2017mean}); these refinements are standard in semi-supervised image classification but unnecessary here, given the small patch size ($16 \times 16$) and the stronger cross-variable signal from $\mathcal{L}_\text{phys}$. The augmented forward pass skips the imputation heads, BiasHead and physics module since they do not contribute to $\mathcal{L}_\text{cons}$.

\paragraph{Imputation training loss.}
\begin{equation}
    \mathcal{L}_\text{imp} = \frac{1}{\sum_{k,i,j} R_{k,i,j}}
    \sum_{k,i,j} R_{k,i,j} \bigl(\hat{m}_{k,i,j} - y_{k,i,j}\bigr)^2.
    \label{eq:imp_loss}
\end{equation}
This is naive masked MSE between the imputation head outputs $\hat{m}_{k,i,j}$ and the labels $y_{k,i,j}$, applied only at labelled pixels. The imputation heads are trained \emph{exclusively} through this loss: gradients from $\mathcal{L}_\text{sup}$ are stopped at $\hat{m}$ (cf.\ equation~(\ref{eq:dr_pseudooutcome})), so the outcome estimator and the regression head do not share gradient flow. This preserves the independence between the outcome model and the propensity correction that gives AIPW its variance-reduction property over standard IPW. The imputation heads are themselves biased toward the observed distribution (since they are trained on labelled pixels only); this is the standard AIPW design, and the bias is corrected by the propensity weighting in $\mathcal{L}_\text{sup}$. The architecture of the imputation heads matches the regression heads (Appendix~\ref{app:arch}), giving the imputation model the same representational capacity as the deployed regression model.

\section{Source-balanced sampling}
\label{app:Source-balanced sampling}

Without intervention, a uniform sampler over $\mathcal{D} = \mathcal{D}_G \cup \mathcal{D}_P$ would draw samples in proportion to source size: $\sim$99.6\% GEDI / 0.4\% plot for Africa, $\sim$99.1\% / 0.9\% for Spain. The plot-supervised heads (AGB, SD, WD) would then receive gradient updates in fewer than 1\% of batches, and the physics consistency loss, which requires GEDI-calibrated $H, C$ and plot-calibrated AGB to coexist in a single batch, would rarely receive a balanced cross-source pairing. As a result, we use a source-balanced sampler where every batch is constructed as a 50/50 mix:
\begin{equation}
    \text{Batch} \;=\; \underbrace{\mathbf{x}_1^G, \ldots, \mathbf{x}_{B/2}^G}_{\text{GEDI}}
    \;\cup\; \underbrace{\mathbf{x}_1^P, \ldots, \mathbf{x}_{B/2}^P}_{\text{Plot}}.
\end{equation}
GEDI samples are shuffled without replacement per epoch; plot samples are oversampled with replacement. Epoch length is $\lfloor N_G / (B/2) \rfloor$, so each GEDI sample is seen approximately once per epoch and each plot sample is seen $N_G/N_P$ times.

\paragraph{Effective passes per epoch.}
Because plot data are heavily oversampled, the two sources converge on different timescales:

\begin{table}[h]
\centering
\small
\begin{tabular}{lccc}
\toprule
& GEDI ($H, C$) & Plot, Spain & Plot, Africa \\
\midrule
Training samples & $\sim$2{,}000{,}000 & $\sim$20{,}000 & $\sim$8{,}000 \\
Test samples & $\sim$10{,}000 & $\sim$6{,}000 & 162 \\
Effective passes per epoch & 1 & $\sim$110 & $\sim$250 \\
Physics warmup bottleneck? & yes & no & no \\
\bottomrule
\end{tabular}
\end{table}

\noindent The physics warmup is therefore governed by epoch count (which tracks GEDI convergence rather than step count or plot pass count).

\paragraph{Test set size.}
The test set sizes (Spain: $\sim$6{,}000 plots covering the region of interest; Africa: 162 standardised 1\,ha plots, recommended by the SEOSAW data owners as the standard and higher-quality plots) further motivate the asymmetric experimental treatment in section~\ref{sec:results}: stratified analyses are reported on Spain, while Africa supports aggregate-level claims about portability.

\paragraph{Why 50/50.}
Lower plot ratios under-represent the only source of AGB supervision; higher plot ratios under-represent GEDI and degrade the structural predictions that feed $\mathcal{L}_\text{phys}$. A symmetric ratio gives the encoder balanced gradient from both label structures every step, and is the ratio used throughout.

\paragraph{Source-balanced sampling does not affect propensity calibration.}
The source-balanced sampler operates at the patch level, balancing how often GEDI patches and plot patches enter a batch. The propensity $\hat{\boldsymbol{\pi}}$ and the AIPW correction operate at the pixel level within a patch, modelling whether a given variable $k$ at pixel $(i,j)$ is observed conditional on the EO covariates $\mathbf{x}$. These two distributions are independent: changing how often patches of each source enter a batch does not change the within-patch distribution of which pixels carry plot labels, which is what the BiasHead is trained to predict. The MNAR mechanism we correct for is therefore the surveyor accessibility bias inside each plot-supervised patch, not the source imbalance between GEDI and plot patches.

\section{Implementation and training}
\label{app:impl}

The model is implemented in PyTorch~\citep{paszke2019pytorch} and trained with AdamW~\citep{loshchilov2017decoupled}, with peak learning rate 5.0e-4 and weight decay 1.0e-4. The learning-rate schedule is cosine annealing with linear warmup over the first 100{,}000 training steps. We use a batch size of $B = 32$ (split 50/50 between GEDI and plot samples by the source-balanced sampler; see Appendix~\ref{app:Source-balanced sampling}). One epoch corresponds to $\lfloor N_G / (B/2) \rfloor \approx 125,000$ steps.

We validate every 500 training steps; the best checkpoint is selected by AGB RMSE on a hold out validation set, with early stopping patience of 250 validation checks ($\approx$ one epoch). Training runs for a maximum of 100 epochs ($\sim$12{,}500{,}000 steps); typical convergence is reached within 50 epochs. Full StruMPL training takes approximately 40 hours per site on a single NVIDIA L4 GPU. The architecture, loss components, hyperparameters, and training recipe are identical across the two sites; only the choice of $g_{\boldsymbol{\phi}}$ varies (with vs.\ without WD).

The forward pass conditionally skips the physics, imputation and propensity modules when their loss weights are zero, so ablations run at the cost of a plain multi-task regressor without any code branching. 

\paragraph{Loss weight values.}
We use $\lambda_\text{phys}^\text{start} = 0.05$, $\lambda_\text{phys}^\text{end} = 0.1$, $T_\text{phys} = 20$\,epochs, $\lambda_\text{cons} = 0.1$, $\lambda_\text{bias} = 0.1$, and $\lambda_\text{imp} = 1.0$. These were selected on a hold-out validation split via a coarse grid search over each weight independently while keeping the others fixed at their default value.

\paragraph{Numerical stability.}
The propensity is clamped to $[\pi_\text{min}, 1]$ with $\pi_\text{min} = 0.1$ to bound the maximum IPW weight at $10\times$. The physics module is computed in log-space with the structural exponent $\text{sp}(\hat y_\text{SD}) \cdot d$ clamped to $[-10, 10]$ to prevent overflow on early predictions. The physics output is clamped to the ecologically plausible range $[0, 2000]$\,Mg/ha before renormalisation. All raw exponent parameters in the physics module are initialised to $-4$, giving $\text{sp}(-4) \approx 0.018$ post-softplus and an initial $\widehat{\text{AGB}}_\text{phys}$ in the range 20--25\,Mg/ha regardless of input magnitudes.

\paragraph{Reproducibility.}
We report headline numbers as mean $\pm$ standard deviation over 5 random seeds (42, 123, 456, 789, 1011); ablation rows are single-seed (42). All sources of randomness (PyTorch, NumPy, Python \texttt{random}, and CUDA operations) are seeded in each run.

\section{Stop-gradient analysis}
\label{app:gradient}

We provide complete gradient calculations for the supervised loss $\mathcal{L}_\text{sup}$ (Equation~\eqref{eq:sup_loss}) under three stop-gradient configurations, demonstrating why both detachments are individually necessary. All gradients are computed pixelwise; the constant factor $\frac{2}{KHW}$ is absorbed throughout.

\paragraph{Setup.}
The supervised loss is $\mathcal{L}_\text{sup} = \frac{1}{KHW} \sum (\hat{y} - \tilde{y})^2$ with pseudo-outcome $\tilde{y} = \mu + \frac{R}{\hat{\pi}}(y - \mu)$. At unlabelled pixels ($R=0$): $\tilde{y} = \mu$. At labelled pixels ($R=1$): $\tilde{y} = \mu + \frac{1}{\hat{\pi}}(y - \mu)$.

\paragraph{Case 1: both $\mu$ and $\hat{\pi}$ detached (default).}
With $\mu$ and $\hat{\pi}$ treated as constants, the only gradient flow is into $\hat{y}$. At labelled pixels: 

$$\frac{\partial \mathcal{L}_\text{sup}}{\partial \hat{y}} \propto 2\left(\hat{y} - \mu - \frac{1}{\hat{\pi}}(y - \mu)\right).$$

At unlabelled pixels: 

$$\frac{\partial \mathcal{L}_\text{sup}}{\partial \hat{y}} \propto 2(\hat{y} - \mu).$$

The unlabelled-pixel term provides AIPW's signal where labels are absent: $\hat{y}$ is pulled toward the imputation prediction $\mu$, a control-variate signal that does not exist in standard IPW. The imputation $\hat{m}$ and propensity $\hat{\pi}$ are unchanged by $\mathcal{L}_\text{sup}$ and trained only through their respective losses ($\mathcal{L}_\text{imp}$, $\mathcal{L}_\text{bias}$).

\paragraph{Case 2: only $\mu$ detached, $\hat{\pi}$ has gradient.}
At labelled pixels, the gradient w.r.t.\ $\hat{\pi}$ is non-zero. Using $\frac{\partial \tilde{y}}{\partial \hat{\pi}} = -\frac{1}{\hat{\pi}^2}(y - \mu)$: 

$$\frac{\partial \mathcal{L}_\text{sup}}{\partial \hat{\pi}} \propto \frac{2}{\hat{\pi}^2}(\hat{y} - \tilde{y})(y - \mu).$$

Substituting $\hat{y} - \tilde{y} = -\frac{1}{\hat{\pi}}(y - \mu)$ at the linearization $\hat{y} \approx \mu$:

$$\frac{\partial \mathcal{L}_\text{sup}}{\partial \hat{\pi}} \propto -\frac{2}{\hat{\pi}^3}(y - \mu)^2 < 0.$$

This negative gradient on $\hat{\pi}$ causes gradient descent to \emph{increase} $\hat{\pi}$, eventually saturating at $\hat{\pi} \to 1$. Once $\hat{\pi} \approx 1$, the IPW correction $1/\hat{\pi}$ has no effect and $\tilde{y} = y$ at labelled pixels, reducing $\mathcal{L}_\text{sup}$ to naive masked MSE. This is the propensity-collapse failure mode empirically observed in Table~\ref{tab:detach}.

\paragraph{Case 3: only $\hat{\pi}$ detached, $\mu$ has gradient.}
At labelled pixels: $\frac{\partial \tilde{y}}{\partial \hat{m}} = (1 - \frac{1}{\hat{\pi}})$. So:

$$\frac{\partial \mathcal{L}_\text{sup}}{\partial \hat{m}} \propto -2(\hat{y} - \tilde{y})(1 - \tfrac{1}{\hat{\pi}}) = 2(\hat{y} - \tilde{y})(\tfrac{1}{\hat{\pi}} - 1).$$

At unlabelled pixels: $\frac{\partial \mathcal{L}_\text{sup}}{\partial \hat{m}} \propto -2(\hat{y} - \mu)$. Both pull the imputation $\hat{m}$ toward the regression head $\hat{y}$, collapsing $\hat{m} \to \hat{y}$. With $\hat{m} \approx \hat{y}$, the supervised loss at labelled pixels becomes:

$$\mathcal{L}_\text{sup}^{R=1} = \left(\hat{y} - \tilde{y}\right)^2 = \left(-\tfrac{1}{\hat{\pi}}(y - \hat{y})\right)^2 = \tfrac{1}{\hat{\pi}^2}(y - \hat{y})^2.$$

This is a $1/\hat{\pi}^2$-weighted MSE (strictly worse than standard IPW's $1/\hat{\pi}$ weighting). Combined with the loss of the AIPW unlabelled-pixel signal (which now contributes zero gradient in expectation as $\hat{y}$ matches $\mu$), the framework reduces to unstable IPW. This is the imputation-collapse failure mode empirically observed in Table~\ref{tab:detach}.

\paragraph{Summary.}
Both stop-gradients are individually necessary: detaching $\hat{\pi}$ prevents the encoder from inflating propensities to escape the MNAR correction; detaching $\mu$ preserves the independence between $\hat{m}$ and $\hat{y}$ that gives AIPW its control-variate property. Either omission collapses the framework to a degenerate variant of standard MSE or IPW.

\section{External baseline evaluation}
\label{app:external}

We compare against two external baselines:

\paragraph{Closest published method~\citep{asiyabi2026process}.}
The~\citep{asiyabi2026process} method was implemented from the published description and trained from scratch on identical training and test splits to those used for StruMPL and the ablations on each site (SNFI for Spain, SEOSAW for Africa). Input channels ($C_{in} = 15$), test plot selection, AGB label handling, and evaluation metrics are identical to those used elsewhere in this paper. The training procedure and hyperparameters follow the~\citep{asiyabi2026process} paper. The reported RMSE and bias in Table~\ref{tab:main} are computed on the same test pixels as full StruMPL, enabling a fair comparison. This is the closest published method to StruMPL in problem setting and is the more informative of the two external baselines.

\paragraph{Off-the-shelf standard reference~\citep{santoro2024biomasscci} (ESA CCI Biomass v5.01).}
The European Space Agency (ESA) Climate Change Initiative (CCI) Biomass map is a global gridded AGB product. Unlike the~\citep{asiyabi2026process} baseline, this is a published map rather than a method we re-ran: we extract the ESA CCI AGB values at the geographic locations of our test plots in both sites and compare them to the corresponding labels from field-plots using the same RMSE and bias metrics as for the trained models. The ESA CCI map is included as a reference point for how an off-the-shelf operational global product compares to site-specific trained models; it is not a like-for-like baseline because it was not trained or calibrated on either of our datasets, and its evaluation is not adjusted for any spatial-resolution or temporal mismatch between the map and the test plots. We report it for transparency and as a benchmark for the order-of-magnitude error a non-customised product delivers, not as a method-level comparison.

\section{Physics module variants}
\label{app:phys_variants}

The form in equation~\eqref{eq:allometric_phys} is one of several differentiable parameterisations we considered. We summarise the alternatives below. All are computed in log-space, all use $\text{sp}(\cdot)+\varepsilon$ for stability, and all have parameters constrained positive via softplus or exp.

\begin{table}[h]
\centering
\small
\begin{tabular}{lll}
\toprule
Variant & Form & Notes \\
\midrule
\texttt{power\_law} & $\widehat{\text{AGB}} = \text{scale} \cdot \prod_i \text{sp}(x_i)^{e_i}$ & $n$ inputs; baseline; $n+1$ params \\
\texttt{allometric} (default, this paper) & $\alpha + \text{scale} \cdot [H^b C^c]^{\text{SD}\cdot d} \cdot \text{WD}^e$ & 6 params; needs WD \\
\texttt{allometric\_no\_wd} & $\alpha + \text{scale} \cdot [H^b C^c]^{\text{SD}\cdot d}$ & 5 params; Spain default \\
\texttt{mlp} & $\widehat{\text{AGB}} = \text{MLP}([x_1, \ldots, x_n])$ & expressive but not interpretable \\
\bottomrule
\end{tabular}
\end{table}

\noindent We compare \texttt{allometric} against \texttt{power\_law} and \texttt{mlp} as ablations in section~\ref{sec:results}. \texttt{allometric\_no\_wd} is structurally \texttt{allometric} without the WD factor and is the default at SNFI; we use the same $-4$ initialisation strategy across all variants (except for MLP which is initialised randomly upon model definition).

\section{Statistical significance of the Spain RMSE improvement}
\label{app:bootstrap}

The headline RMSE improvement of full StruMPL over the strongest ablation on Spain (41.9 vs.\ 43.0\,Mg/ha for MTL + $\mathcal{L}_\text{bias} + \mathcal{L}_\text{imp}$) is small relative to the seed-to-seed standard deviation ($\pm$0.5\,Mg/ha across 5 seeds). To check whether the gap is statistically robust or within sampling noise, we run a paired bootstrap on the Spain test set.

\paragraph{Procedure.}
For each of $n = 10{,}000$ bootstrap iterations, we resample the test pixels with replacement (preserving the labelled-pixel distribution and the original sample size) and compute AGB RMSE for both configurations on the resample. The difference $\Delta = \text{RMSE}_{\text{StruMPL}} -  \text{RMSE}_{\text{ablation}}$ is recorded for each iteration. Because both configurations are evaluated on identical resampled pixels, the test is paired: variability in which samples are selected cancels out, and only the model difference contributes to the variance of $\Delta$.

\paragraph{Result.}
The observed RMSE difference is -1.1\,Mg/ha (full StruMPL lower). The 95\% bootstrap CI on $\Delta$ is [-1.598, -0.522]\,Mg/ha, entirely below zero. The one-sided p-value (probability of observing $\Delta \geq 0$ under the bootstrap distribution) is $p < 0.05$. We therefore claim the Spain RMSE improvement is statistically significant at the 5\% level. The bootstrap shows that full StruMPL's lower aggregate RMSE on Spain is not an artefact of the test sampling. 

\paragraph{Why only Spain.}
The same test is not informative on Africa because the SEOSAW test set contains only 162 plots. At this scale, bootstrap CIs on $\Delta$ are wide enough to admit either sign of the difference for most ablation comparisons, regardless of the underlying truth. We report the Africa point estimates and their 5-seed standard deviations (Table~\ref{tab:full_ablation_africa}) and rely on the larger Spanish test set
for the significance claim.

\section{Full ablation table}
\label{app:full_ablation_table}

Table~\ref{tab:main} and Figure~\ref{fig:RMSE_Bias} in the main paper report a curated subset of configurations. The complete ablation grid, including single-output baselines, all loss-component combinations, different physic model, naive MSE vs. IPW vs.\ AIPW, and stop-gradient ablation comparison, is given in Table~\ref{tab:full_ablation_spain} for all AGB and other structural variables for SNFI and in Table~\ref{tab:full_ablation_africa} for SEOSAW datasets.

\begin{table}[h]
\centering
\footnotesize
\caption{Full ablation grid on the SNFI (Spain) test set. AGB RMSE in Mg/ha; structural variables in their native units.}
\label{tab:full_ablation_spain}
\begin{tabular}{lccccc}
\toprule
Configuration & AGB RMSE & AGB bias & H RMSE & C RMSE & SD RMSE \\
\midrule
\multicolumn{6}{l}{\emph{Single-output baselines}} \\
Single-output AGB, naive MSE & \texttt{43.7}& \texttt{0.9}& --- & --- & --- \\
Single-output AGB + $\mathcal{L}_\text{bias}$ + $\mathcal{L}_\text{cons}$ + $\mathcal{L}_\text{imp}$ & \texttt{43.1}& \texttt{2.1}& --- & --- & --- \\
\midrule
\multicolumn{6}{l}{\emph{Joint multi-task variants (no MNAR correction, no physics)}} \\
MTL, sup only & \texttt{43.9}& \texttt{1.3}& \texttt{3.1}& \texttt{0.128}& \texttt{459.9}\\
\midrule
\multicolumn{6}{l}{\emph{Loss component additions}} \\
MTL + $\mathcal{L}_\text{phys}$ & \texttt{43.7}& \texttt{0.9}& \texttt{3.0}& \texttt{0.126}& \texttt{459.2}\\
MTL + $\mathcal{L}_\text{bias}$ + $\mathcal{L}_\text{imp}$ & \texttt{43.0}& \texttt{1.5}& \texttt{2.9}& \texttt{0.124}& \texttt{461.4}\\
MTL + $\mathcal{L}_\text{cons}$ & \texttt{43.3}& \texttt{0.8}& \texttt{2.9}& \texttt{0.125}& \texttt{457.4}\\
\midrule
\multicolumn{6}{l}{\emph{Supervised loss form (other components held at full StruMPL)}} \\
StruMPL with naive masked MSE & \texttt{43.2}& \texttt{1.7}& \texttt{3.9}& \texttt{0.125}& \texttt{468.4}\\
StruMPL with IPW supervised loss & \texttt{44.3}& \texttt{0.7}& \texttt{3.1}& \texttt{0.126}& \texttt{469.3}\\
\raisebox{1\height}{\textbf{StruMPL with AIPW (default)}} &
  \shortstack[c]{\texttt{41.9}\\[1pt] \texttt{$\pm$ 0.5}} &
  \shortstack[c]{\texttt{0.4}\\[1pt] \texttt{$\pm$ 0.1}} &
  \shortstack[c]{\texttt{2.8}\\[1pt] \texttt{$\pm$ 0.05}} &
  \shortstack[c]{\texttt{0.123}\\[1pt] \texttt{$\pm$ 0.001}} &
  \shortstack[c]{\texttt{457.9}\\[1pt] \texttt{$\pm$ 3.5}} \\
\midrule
\multicolumn{6}{l}{\emph{Physics module form (other components held at full StruMPL)}} \\
StruMPL with \texttt{power\_law} physics & \texttt{44.3}& \texttt{1.0}& \texttt{3.3}& \texttt{0.125}& \texttt{473.2}\\
StruMPL with \texttt{mlp} physics & \texttt{44.4}& \texttt{1.1}& \texttt{3.0}& \texttt{0.126}& \texttt{468.7}\\
\midrule
\multicolumn{6}{l}{\emph{Stop-gradient ablation (default = both detached)}} \\
StruMPL, $\hat\pi$ \emph{not} detached & \texttt{44.2}& \texttt{1.8}& \texttt{3.9}& \texttt{0.127}& \texttt{469.2}\\
StruMPL, $\mu = \hat y$ \emph{not} detached & \texttt{44.6}& \texttt{1.5}& \texttt{4.8}& \texttt{0.123}& \texttt{469.2}\\
\bottomrule
\end{tabular}
\end{table}

\begin{table}[h]
\centering
\scriptsize
\caption{Full ablation grid on the SEOSAW (Africa) test set. AGB RMSE in Mg/ha; structural variables in their native units.}
\label{tab:full_ablation_africa}
\begin{tabular}{lcccccc}
\toprule
Configuration & AGB RMSE & AGB bias & H RMSE & C RMSE & SD RMSE & WD RMSE \\
\midrule
\multicolumn{7}{l}{\emph{Single-output baselines}} \\
Single-output AGB, naive MSE & \texttt{22.5}& \texttt{5.3}& --- & --- & --- & --- \\
Single-output AGB + $\mathcal{L}_\text{bias}$ + $\mathcal{L}_\text{cons}$ + $\mathcal{L}_\text{imp}$ & \texttt{22.6}& \texttt{2.9}& --- & --- & --- & --- \\
\midrule
\multicolumn{7}{l}{\emph{Joint multi-task variants (no MNAR correction, no physics)}} \\
MTL, sup only & \texttt{24.9}& \texttt{1.4}& \texttt{3.9}& \texttt{0.153}& \texttt{97.9}& \texttt{0.111}\\
\midrule
\multicolumn{7}{l}{\emph{Loss component additions}} \\
MTL + $\mathcal{L}_\text{phys}$ & \texttt{24.1}& \texttt{5.4}& \texttt{3.9}& \texttt{0.152}& \texttt{92.6}& \texttt{0.112}\\
MTL + $\mathcal{L}_\text{bias}$ + $\mathcal{L}_\text{imp}$ & \texttt{23.6}& \texttt{2.2}& \texttt{4.1}& \texttt{0.160}& \texttt{98.4}& \texttt{0.111}\\
MTL + $\mathcal{L}_\text{cons}$ & \texttt{23.7}& \texttt{1.3}& \texttt{4.1}& \texttt{0.158}& \texttt{108.6}& \texttt{0.113}\\
\midrule
\multicolumn{7}{l}{\emph{Supervised loss form (other components held at full StruMPL)}} \\
StruMPL with naive masked MSE & \texttt{25.3}& \texttt{1.7}& \texttt{3.9}& \texttt{0.158}& \texttt{94.1}& \texttt{0.115}\\
StruMPL with IPW supervised loss & \texttt{24.1}& \texttt{2.6}& \texttt{3.9}& \texttt{0.157}& \texttt{97.8}& \texttt{0.114}\\
\raisebox{1\height}{\textbf{StruMPL with AIPW (default)}} &
  \shortstack[c]{\texttt{20.7}\\[1pt] \texttt{$\pm$ 0.6}}&
  \shortstack[c]{\texttt{0.6}\\[1pt] \texttt{$\pm$ 0.7}}&
  \shortstack[c]{\texttt{3.8}\\[1pt] \texttt{$\pm$ 0.1}}&
  \shortstack[c]{\texttt{0.153}\\[1pt] \texttt{$\pm$ 0.003}}&
  \shortstack[c]{\texttt{90.5}\\[1pt] \texttt{$\pm$ 3.8}}&
  \shortstack[c]{\texttt{0.110}\\[1pt] \texttt{$\pm$ 0.007}}\\
\midrule
\multicolumn{7}{l}{\emph{Physics module form (other components held at full StruMPL)}} \\
StruMPL with \texttt{power\_law} physics & \texttt{25.5}& \texttt{7.3}& \texttt{3.8}& \texttt{0.151}& \texttt{112.1}& \texttt{0.122}\\
StruMPL with \texttt{mlp} physics & \texttt{24.2}& \texttt{3.4}& \texttt{3.9}& \texttt{0.156}& \texttt{98.6}& \texttt{0.119}\\
\midrule
\multicolumn{7}{l}{\emph{Stop-gradient ablation (default = both detached)}} \\
StruMPL, $\hat\pi$ \emph{not} detached & \texttt{25.0}& \texttt{3.9}& \texttt{3.8}& \texttt{0.155}& \texttt{92.5}& \texttt{0.112}\\
StruMPL, $\mu = \hat y$ \emph{not} detached & \texttt{24.8}& \texttt{3.6}& \texttt{3.6}& \texttt{0.154}& \texttt{95.7}& \texttt{0.128}\\
\bottomrule
\end{tabular}
\end{table}

\section{Propensity calibration}
\label{app:calibration}

\begin{figure}[h]
    \centering
    \includegraphics[width=0.5\textwidth]{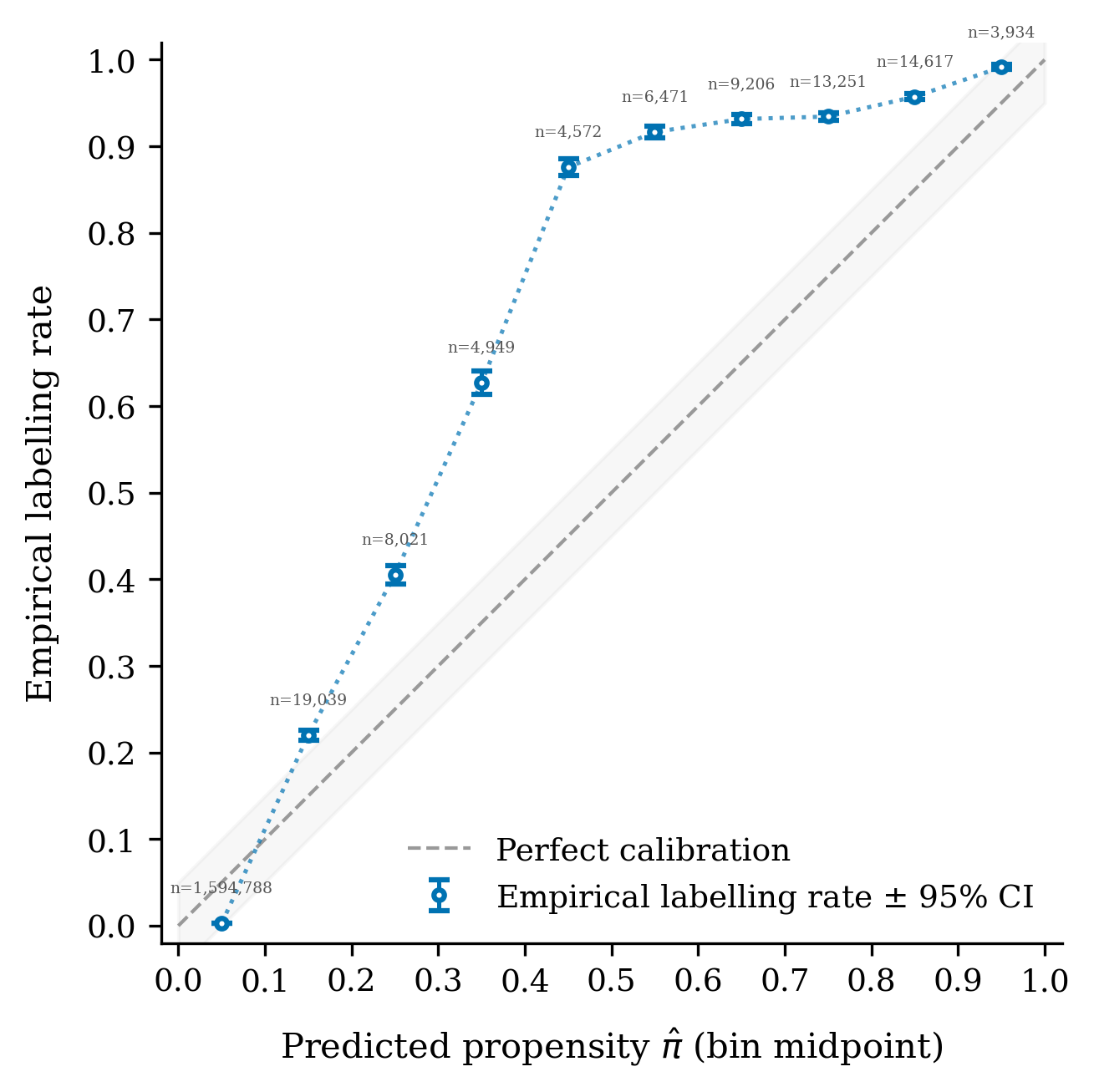}
    \caption{Propensity calibration on the Spain test set. Predicted $\hat\pi$ is binned into deciles; the empirical labelling rate within each bin is plotted against the bin midpoint, with the diagonal indicating perfect calibration. Calibration is good at both ends of the range ($\hat\pi \leq 0.15$ and $\hat\pi \geq 0.85$) and weaker in the mid-range, where the model under-predicts the true labelling rate. The monotonic ordering across all bins supports the use of $\hat\pi$ as a ranking signal in the AIPW correction.}
    \label{fig:calibration}
\end{figure}

The AIPW correction relies on $\hat\pi$ approximating the true propensity. We diagnose the propensity head's calibration on Spain dataset by binning test pixels by predicted $\hat\pi$ and computing the empirical labelling rate within each bin (Figure~\ref{fig:calibration}).

Figure~\ref{fig:calibration} shows the predicted propensity $\hat\pi$ binned into deciles against the empirical labelling rate within each bin, on the Spain test set. Three patterns are worth noting. First, the propensity head is \emph{monotonically calibrated}: the empirical labelling rate increases with $\hat\pi$ across all ten bins, from 0.003 at $\hat\pi \approx 0.05$ to 0.991 at $\hat\pi \approx 0.95$. This correct ranking is the property AIPW relies on most directly as the relative ordering of weights determines whether the bias correction re-weights the right pixels. Second, calibration is best at the extremes ($\hat\pi \in \{0.05, 0.15, 0.85, 0.95\}$) and weakest in the middle range, where the model under-predicts the true labelling rate (e.g., $\hat\pi \approx 0.45$ vs.\ empirical rate $\approx 0.88$). The practical impact of this mid-range miscalibration is bounded by the propensity clamp $\pi_\text{min} = 0.1$, which caps the maximum IPW weight at $10\times$ regardless of the true rate; mid-range pixels carry IPW weights between $\sim$1 and 2, where small calibration errors lead to small weight errors. Third, the BiasHead produces non-trivial spatial variation in $\hat\pi$ within a single batch: the per-pixel standard deviation of $\hat\pi$ within a typical test batch is $\sim$0.13, meaning that the model does not collapse to the degenerate solution in which the BiasHead collapses to a constant $\hat\pi \approx \bar R$ and AIPW reduces to a global re-scaling of the loss. Together, these observations support the use of $\hat\pi$ as the IPW weight in the AIPW correction, not as a maximum-likelihood estimate of the true propensity (which it is not), but as a calibrated ordering of pixels by labelling probability that allows the correction in section~\ref{sec:stratified} to focus the supervisory signals on the under-represented regimes of the training data.

\section{Cross-source portability}
\label{app:portability}

The Africa configuration is structurally different from Spain in three ways: a different biome (dry tropical/savanna vs.\ Mediterranean), a different $\mathcal{K}_P$ ($\{$SD, AGB, WD$\}$ vs.\ $\{$SD, AGB$\}$), and a different physics function $g$. The architecture, optimiser, loss components, weights, and training schedule are identical.

\paragraph{Discussion.}
The Africa results support the portability claim: the unified mask design adapts to the wood density variable availability, the \textit{BalancedBatchSampler} handles the larger source imbalance ($N_G/N_P \approx 250$ vs.\ Spain's $\approx 110$) without recipe changes, and the curriculum warmup remains effective with a different physics functional form. We do not claim biome-stratified bias correction at Africa given the test set size (162 standardised 1\,ha plots are selected on data-provider recommendation as the most reliable subset of the network).

\section{Effect of curriculum warmup}
\label{app:warmup}

The physics-loss uses a curriculum schedule to increase $\lambda_\text{phys}$ as the output of the regression head becomes meaningful. Without warmup ($\lambda_\text{phys}$ at $\lambda_\text{phys}^\text{end}$ from step 0), the physics consistency loss penalises the inconsistency of near-random predictions during the first few thousand steps, contributing noise to the gradient. We observe that without this warmup phase, the model convergence is delayed and the final RMSE is higher.

\section{When the StruMPL formulation applies}
\label{app:generality}

The problem formulation in section~\ref{sec:problem} is domain-neutral: it refers to inputs $\mathbf{x}$, targets $\mathbf{y}$, an observation mask $R$, source-determined label subsets $\mathcal{K}_s$, and a known constraint $g_{\boldsymbol{\phi}}$. The forest-attribute application (section~\ref{sec:problem}, last paragraph) is one example. To clarify the scope of the framework and make the generality claim testable, we list four structural conditions that determine whether StruMPL is the right tool for a given problem. A domain in which all four hold is a candidate.

\paragraph{Condition 1: source-determined disjoint partial supervision.} Training data come from two or more measurement sources, and the source determines which targets are labelled: $\mathcal{K}_s \cap \mathcal{K}_{s'} = \emptyset$ for $s \neq s'$, with no single sample observing all targets. This rules out standard partial-label settings where any subset of labels may be missing per sample.

\paragraph{Condition 2: missingness depends on the latent label.} Within at least one source, the labelled subset of locations is not a random sample of the population: $P(R=1 \mid \mathbf{x}, y) \neq P(R=1 \mid \mathbf{x})$ in some part of the input space.

\paragraph{Condition 3: known structural constraint among targets.} A function $g_{\boldsymbol{\phi}}$ that links a derived target to other targets is given by domain knowledge, with parameters $\boldsymbol{\phi}$ either known or learnable. The constraint must be differentiable in the target predictions; for instance: closed-form algebraic relationships, algebraic conservation laws, and parameterised empirical laws. However, pure black-box constraints (e.g., predictions of a separately trained model) are not what the physics module is designed for.

\paragraph{Condition 4: dense or otherwise structured prediction.} The target is predicted at many locations per sample (pixels, time steps, mesh nodes), so the propensity field $\hat{\boldsymbol{\pi}}$ and the per-location physics residual carry information beyond a scalar. For purely scalar regression, the AIPW correction is still meaningful but the encoder-shared propensity field reduces to a sample-level scalar and the per-location physics constraint becomes a single residual per sample, undermining most of the framework's value.

\paragraph{Candidate application areas.} Several settings outside Earth observation plausibly satisfy all four conditions:

\textit{Hydrology.} Streamflow, evapotranspiration, soil moisture, and snow water equivalent are observed by different sensor networks (USGS gauges, FLUXNET towers, SMAP, MODIS, in-situ snow pillows), each covering different catchments and seasons. Snow measurements in particular are MNAR (concentrated where access is feasible). The water balance $P = ET + Q + \Delta S$ provides a hard structural constraint among the targets. 

\textit{Battery state estimation.} State-of-charge, state-of-health (SoH), capacity, internal resistance, and temperature are observed under different test protocols, with capacity tests typically run only on cells that have not yet failed (an MNAR mechanism on the latent SoH). Coulomb counting and impedance models provide structural relationships among the targets.

\textit{Multi-modal medical imaging.} Different imaging modalities (MRI sequences, CT, PET) are acquired in different patient subsets, with acquisition strongly correlated with disease severity (MNAR). Anatomical-consistency constraints across modalities (though softer than a closed-form law) give a candidate $g_{\boldsymbol{\phi}}$. However, we note that the constraint condition is weaker than the other candidate cases, since anatomical consistency is empirical rather than physical.

\paragraph{Where StruMPL is not the right tool.} If labels are randomly missing rather than source-partitioned, a standard partial-label MTL method is sufficient. If observation is MAR rather than MNAR (true for many citizen-science and crowdsourced datasets after geographic stratification), IPW reweighting is unnecessary. If no inter-target structural law is known, the physics module degenerates to a learned regression head with no benefit over additional supervised heads. The framework is targeted at a specific intersection of conditions, not a generic multi-task learner.

\end{document}